\documentclass[runningheads]{llncs}

\usepackage{eccv}

\usepackage{eccvabbrv}

\usepackage{graphicx}
\usepackage{booktabs}
\usepackage{amsmath}
\usepackage{multirow}
\usepackage{float}
\usepackage{wrapfig}
\usepackage{bm}
\usepackage[misc]{ifsym}

\usepackage[accsupp]{axessibility}  % Improves PDF readability for those with disabilities.

\usepackage{hyperref}

\usepackage{orcidlink}

\begin{document}

% ---------------------------------------------------------------
% TODO REVIEW: Replace with your title
\title{Implicit Filtering for Learning Neural Signed Distance Functions from 3D Point Clouds} 

% TODO REVIEW: If the paper title is too long for the running head, you can set
% an abbreviated paper title here. If not, comment out.
\titlerunning{Implicit Filtering}

% TODO FINAL: Replace with your author list. 
% Include the authors' OCRID for the camera-ready version, if at all possible.
\author{Shengtao Li\inst{1,2} \and
Ge Gao\inst{1,2}\textsuperscript{(\Letter)} \and Yudong Liu\inst{1,2} \and Ming Gu\inst{1,2} \and Yu-Shen Liu\inst{2}}

% TODO FINAL: Replace with an abbreviated list of authors.
\authorrunning{S. Li et al.}
% First names are abbreviated in the running head.
% If there are more than two authors, 'et al.' is used.

% TODO FINAL: Replace with your institution list.
\institute{Beijing National Research Center for Information Science and Technology (BNRist), Tsinghua University, Beijing, China\\
\and
School of Software, Tsinghua University, Beijing, China\\
\email{list21@mails.tsinghua.edu.cn, gaoge@tsinghua.edu.cn, liuyd23@mails.tsinghua.edu.cn, guming@tsinghua.edu.cn, liuyushen@tsinghua.edu.cn}
}

\maketitle

\begin{abstract}
  Neural signed distance functions (SDFs) have shown powerful ability in fitting the shape geometry. However, inferring continuous signed distance fields from discrete unoriented point clouds still remains a challenge. The neural network typically fits the shape with a rough surface and omits fine-grained geometric details such as shape edges and corners. In this paper, we propose a novel non-linear implicit filter to smooth the implicit field while preserving high-frequency geometry details. Our novelty lies in that we can filter the surface (zero level set) by the neighbor input points with gradients of the signed distance field. By moving the input raw point clouds along the gradient, our proposed implicit filtering can be extended to non-zero level sets to keep the promise consistency between different level sets, which consequently results in a better regularization of the zero level set. We conduct comprehensive experiments in surface reconstruction from objects and complex scene point clouds, the numerical and visual comparisons demonstrate our improvements over the state-of-the-art methods under the widely used benchmarks. Project page: \url{https://list17.github.io/ImplicitFilter}.
\end{abstract}

\setlength{\intextsep}{2pt}%
\setlength{\columnsep}{8pt}%

\section{Introduction}
\label{sec:intro}
Reconstructing surfaces from 3D point clouds is an important task in 3D computer vision. Recently signed distance functions (SDFs) learned by neural networks have been a widely used strategy for representing high-fidelity 3D geometry. These methods train the neural networks to predict the signed distance for every position in the space by signed distances from ground truth or inferred from the raw 3D point cloud. With the learned signed distance field, we can obtain the surface by running the marching cubes algorithm\cite{marchingcubes} to extract the zero level set.

\begin{figure}[t]
    \centering
    \includegraphics[width=\textwidth]{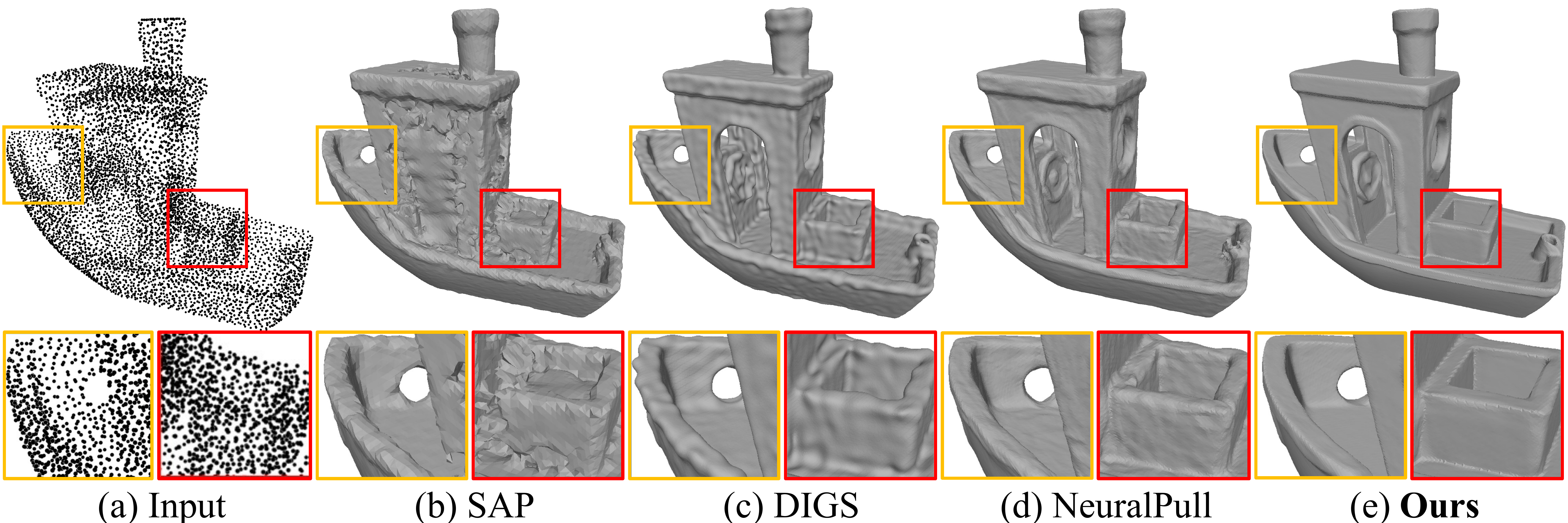}
    \caption{Visualization of the comparisons on FAMOUS dataset\cite{Points2Surf}. Our implicit filter can improve the reconstruction by removing the noise and keeping the geometric details compared with other methods.}
    \label{fig:teaser}
\end{figure}

Without signed distance ground truth, inferring the correct gradient and distance for each query point could be hard. Since the gradient of the neural network also indicates the direction in which the signed distance field changes, recent works\cite{SAL, sald, IGR, siren, BaoruiTowards, NPull} typically add constraints on the network gradient to learn a stable field. In terms of the rate at which the field is changing, the eikonal term\cite{SAL, sald, siren, ben2022digs} is widely used to ensure the norm of the gradient to be one everywhere. For the gradient direction constraint, some methods\cite{NPull, chao2023gridpull} use the direction from the query point to the nearest point on the surface as guidance. Leveraging the continuity of the neural network and the gradient constraint, all these methods could reconstruct discrete points. However, the continuity cannot guarantee the prediction is correct everywhere. Therefore, reconstructed surfaces of previous methods usually contain noise and ignore geometry details when there are not enough points to guide the reconstruction, as shown in \cref{fig:teaser}.

The above issue arises from the fact that these methods overlook the geometric information within the neighborhood but only focus on adding constraints on individual points to optimize the network. To resolve this issue, we introduce the bilateral filter for implicit fields that reduces surface noise while preserving the high-frequency geometric characteristics of the shape. Our designed implicit filter takes into account both the position of point clouds and the gradient of learned implicit fields. Based on the assumption of all input points lying on the surface,  we can filter noise points on the zero level set by minimizing the weighted projection distance to gradients of the neighbor input points. Moreover, by moving the input points along the gradient of the field to other level sets, we can easily extend the filter to the whole field. This helps constrain the signed distance field near the surface and achieve better consistency through different level sets. To evaluate the effectiveness of our proposed implicit filtering, we validate it under widely used benchmarks including object and scene reconstructions. Our contributions are listed below.

\begin{itemize}
    \item We introduce the implicit filter on SDFs to smooth the surface while preserving geometry details for learning better neural networks to represent shapes or scenes.
    \item We improve the implicit filter by extending it to non-zero level sets of signed distance fields. This regularization of the field aligns different level sets and provides better consistency within the whole SDF field.
    \item Both object and scene reconstruction experiments validate our implicit filter, demonstrating its effectiveness and ability to produce high-fidelity reconstruction results, surpassing the previous state-of-the-art methods.
\end{itemize}

\section{Related Work}
With the rapid development of deep learning, neural networks have shown great potential in surface reconstruction from 3D point clouds. In the following, we briefly review methods related to implicit learning for 3D shapes and reconstructions from point clouds.

\noindent \textbf{Implicit Learning from 3D Supervision.}
The most commonly used strategy to train the neural network is to learn priors in a data-driven manner. These methods require signed distances or occupancy labels as 3D supervision to learn global priors \cite{POCO, Points2Surf, mi2020ssrnet, OccupancyNetworks, MLS} or local priors \cite{LIG, Deeplocalshape, patchnets, convonet, saconvonet, alto, li2024GridFormer, huang2023nksr, NKF}. With large-scale training datasets, the neural network can perform well with similar shapes, but may not generalize well to unseen cases with large geometric variations. These models often have limited inputs that can be difficult to scale for varying sizes of point clouds.

\noindent \textbf{Implicit Learning from Raw Point Clouds.}
Different from the supervised methods, we can learn implicit functions by overfitting neural networks on single point clouds globally or locally to learn SDFs \cite{NPull, SAL, sald, chao2023gridpull, SAP, koneputugodage2024smallsteps, PredictiveContextPriors, onsurface, marschner2023constructivesolid}. These unsupervised methods rely on neural networks to infer implicit functions without learning any priors. Therefore, apart from the guidance of original input point clouds, we also need constraints on the direction \cite{NPull, chao2023gridpull, PredictiveContextPriors, koneputugodage2024smallsteps} or the norm \cite{sald, SAL, marschner2023constructivesolid} of the gradients, specially designed priors \cite{PredictiveContextPriors, onsurface}, or differentiable poisson solver \cite{SAP} to infer SDFs. This unsupervised approach heavily depends on the fitting capability and continuity of neural networks. However, these SDFs lack accuracy because there is no reliable guidance available for each query point across the entire space when working with discrete point clouds. Therefore, deducing the correct geometry for free space becomes particularly crucial. Our implicit filtering enhances SDFs by inferring the geometric details through the implicit field information of neighbor points.

\noindent \textbf{Feature Preserving Point Cloud Reconstruction.}
Early works \cite{EAR, FLOP, featurePreserving} reconstruct point clouds with sharp features usually by point cloud consolidation. The key idea of these methods is to enhance the quality of point clouds with sharp features. One popular category is the local projection operation (LOP) \cite{parafree} and its variants \cite{FLOP, conso_unorgan, CLOP, EAR}. The projection operator provides a stable and easily generalizable method for point cloud filtering, which is also the foundation of our implicit filter. The difference lies in that we do not need any normal or other priors and our filtering can be directly applied to implicit fields to extract high-fidelity meshes. Some other learning-based methods \cite{ecnet, pointfilter} try to consolidate point clouds with edge points in a data-driven manner. Although capable of generating high-quality point clouds, these methods still require a proper reconstruction method \cite{RIMLS} to inherit the details in meshes.

With the advancement of deep learning in point cloud reconstruction, some approaches \cite{siren, ben2022digs, edgeimplicit, lindell2022bacon} also explored employing neural networks to reconstruct high-precision models. FFN \cite{FFN}, SIREN \cite{siren}, and IDF \cite{IDF} introduce high-frequency features into the neural network in different ways to preserve the geometric details of the reconstructed shape. DIGS\cite{ben2022digs} and EPI \cite{edgeimplicit} smooth the surface by using the divergence as guidance to alleviate the implicit surface roughness. Compared with these methods, we first introduce local geometric features through filtering to optimize the implicit field, so that we can achieve higher accuracy.

\section{Method}
\subsubsection{Neural SDFs overview.}
This section will briefly describe the concepts we used in our implicit filtering. We focus on the SDF $f : \bm{R}^3 \rightarrow \bm{R}$ inferred from the point cloud $\bm{P} = \{\bm{p}_i | \bm{p}_i \in \bm{R}^3\}_{i=1}^N$ without ground truth signed distances and normals. $f$ predicts a signed distance $s \in \bm{R}$ for an arbitrary query point $\bm{q}$, as formulated by $s = f_\theta(\bm{q})$, where $\theta$ denotes the parameters of the neural network.

The level set $\mathcal{S}_d$ of SDF is defined as a set of continuous query points with the same signed distance $d$, formulated as $\mathcal{S}_d = \{\bm{q}| f_\theta(\bm{q}) = d\}$. The goal of our implicit filtering is to smooth each level set with geometry details. Then we can extract the zero level set as a mesh by running the marching cubes algorithm \cite{marchingcubes}.

% \subsubsection{Level set bilateral filtering.}
\subsubsection{Level set bilateral filtering.}
\begin{wrapfigure}[14]{r}{0pt}
    \includegraphics[width=0.25\textwidth]{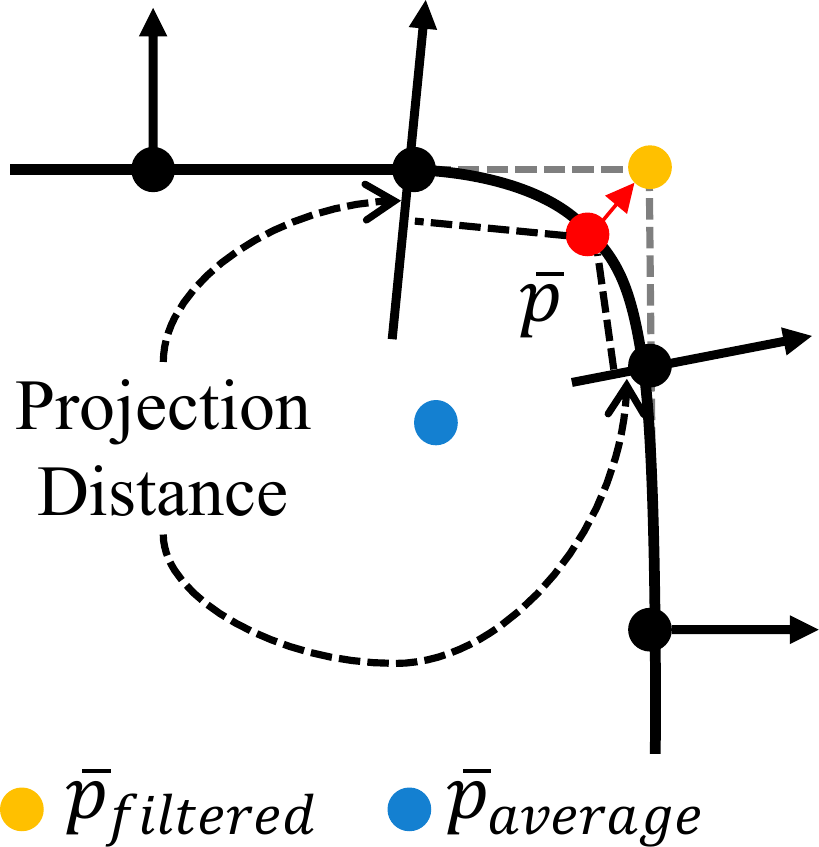}
    \caption{By minimizing the weighted projection distance, our filter can preserve the sharp feature but the average method leads to a wrong result.}
    \label{fig:sharp}
\end{wrapfigure}
Filtering for 2D images replaces the intensity of each pixel with the weighted intensity values from nearby pixels. Different from images, the resolution of implicit fields is infinite and we need to find the neighborhood on each level set for filtering. By minimizing the following loss function, 
\begin{equation}
    L_{dist} = \frac{1}{N}\sum_{i=1}^N|f_\theta(\bm{p}_i)|,
\end{equation}
we can approximate that all points in $\bm{P}$ are located on level set $\mathcal{S}_0$, which makes it feasible to find neighbor points on $\mathcal{S}_0$. For a given point $\bm{\bar{p}}$ on $\mathcal{S_\text{0}}$, one simple strategy of filtering is to average positions of neighbor points $\mathcal{N}(\bm{\bar{p}}, \mathcal{S}_0) \subset \bm{P}$ on $\mathcal{S}_0$ by a Gaussian filter based on relative positions as follows: 

\begin{equation}
\bm{\bar{p}}_{\text{average}} = \frac{\sum_{\bm{p}_j \in \mathcal{N}(\bm{\bar{p}}, \mathcal{S}_0)}{\bm{p}_j \phi(||\bm{\bar{p}} - \bm{p}_j||)}}{{\sum_{\bm{p}_j \in \mathcal{N}(\bm{\bar{p}}, \mathcal{S}_0)}{\phi(||\bm{\bar{p}} - \bm{p}_j||)}}},
\label{eq:averageposition}
\end{equation}
where the Gaussian function $\phi$ is defined as $\phi(||\bm{\bar{p}} - \bm{p}_j||) = \text{exp}\left(-\frac{||\bm{\bar{p}} - \bm{p}_j||^2}{\sigma_p^2}\right).$

\begin{figure*}[t]
    \centering
    \includegraphics[width=\linewidth]{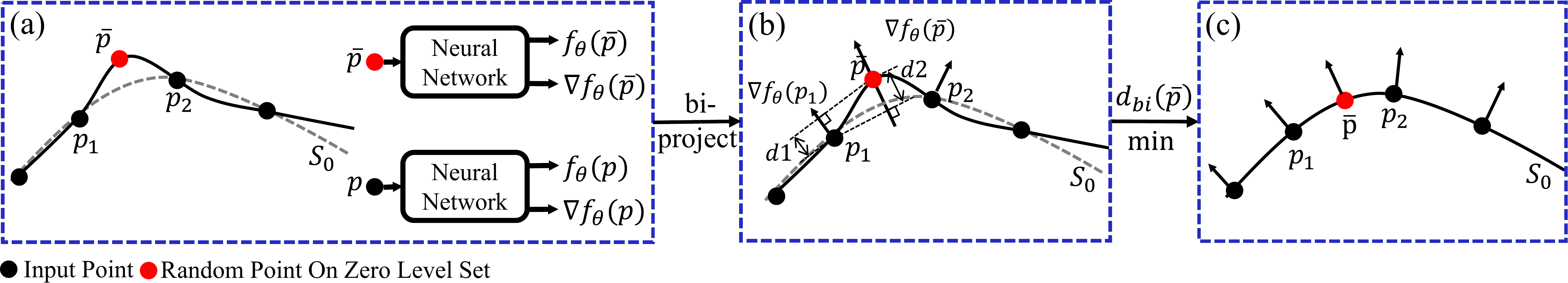}
    \caption{Overview of filtering the zero level set. (a) We assume all input points lying on the surface and compute gradients as normals. (b) Calculating bidirectional projection distances $d1=|\bm{n}_{p_j}^T(\bm{\bar{p}} - \bm{p}_j)|$, $d2 = |\bm{n}_{\bar{p}}^T(\bm{\bar{p}} - \bm{p}_j)|$ and the weights in \cref{eq:dual_project_distance}. (c) By minimizing \cref{eq:dual_project_distance}, we can remove the noise on the zero level set. The gradient $\nabla f_\theta$ in this figure defaults to be regularized.} 
    \label{fig:zerolevelset}
\end{figure*}

However, as depicted in \cref{fig:sharp}, it is evident that this weighted mean position yields excessively smooth surfaces, causing sharp features and details to be further obscured. To keep the geometric details, our filtering operator suggests measuring the projection distance to the gradient of neighbor points as shown in \cref{fig:sharp} and \cref{fig:zerolevelset}(b). When calculating weights, it is vital to account for both the impact of relative positions and the gradient similarity. Following the principles of bilateral filtering, to compute the filtered point for $\bm{\bar{p}}$, we simply need to minimize the following distance equation:

\begin{equation}
\footnotesize
    d(\bm{\bar{p}}) =\frac{\sum_{\bm{p}_j \in \mathcal{N}(\bm{\bar{p}}, \mathcal{S}_0)}{|\bm{n}^T_{p_j}(\bm{\bar{p}} - \bm{p}_j)| \phi(||\bm{\bar{p}} - \bm{p}_j||) \psi(\bm{n}_{\bar{p}}, \bm{n}_{p_j})}}{{\sum_{\bm{p}_j \in \mathcal{N}(\bm{\bar{p}}, \mathcal{S}_0)}{\phi(||\bm{\bar{p}} - \bm{p}_j||) \psi(\bm{n}_{\bar{p}}, \bm{n}_{p_j})}}},
    \label{eq:weighted_project_distance}
\end{equation}
where the gradient $\bm{n}_{\bar{p}}$, $\bm{n}_{p_j}$ and the Gaussian function $\psi$ are defined as $\bm{n}_{\bar{p}} = \frac{\nabla f_\theta(\bm{\bar{p}})}{||\nabla f_\theta(\bm{\bar{p}})||}, \bm{n}_{\bm{p}_j} = \frac{\nabla f_\theta(\bm{p}_j)}{||\nabla f_\theta(\bm{p}_j)||},
    \psi(\bm{n}_{\bar{p}}, \bm{n}_{p_j}) = \text{exp}\left(-\frac{1-\bm{n}_{\bar{p}}^T\bm{n}_{p_j}}{1 - \text{cos}(\sigma_n)}\right).
$

In addition to projection to the gradient $\bm{n}_{p_j}$, we observe that the projection distance to $\bm{n}_{\bar{p}}$ can assist in learning a more stable gradient for point $\bar{p}$ which is also adopted in EAR\cite{EAR}. Taking into account the bidirectional projection, our final bilateral filtering operator can be formulated as follows:

\begin{equation}
\footnotesize
    d_{bi}(\bm{\bar{p}}) =\frac{\sum\limits_{\bm{p}_j \in \mathcal{N}(\bm{\bar{p}}, \mathcal{S}_0)}{\left(|\bm{n}_{p_j}^T(\bm{\bar{p}} - \bm{p}_j)| + |\bm{n}_{\bar{p}}^T(\bm{\bar{p}} - \bm{p}_j)|\right) \phi(||\bm{\bar{p}} - \bm{p}_j||) \psi(\bm{n}_{\bar{p}}, \bm{n}_{p_j})}}{{\sum\limits_{\bm{p}_j \in \mathcal{N}(\bm{\bar{p}}, \mathcal{S}_0)}{\phi(||\bm{\bar{p}} - \bm{p}_j||) \psi(\bm{n}_{\bar{p}}, \bm{n}_{p_j})}}}.
    \label{eq:dual_project_distance}
\end{equation}

Although similar filtering methods have been widely studied in applications such as point cloud denoising and resampling\cite{pointfilter, EAR}, there are two critical problems when applying these methods in implicit fields:
\begin{enumerate}
    \item Filtering the zero level set needs to sample points on the level set $\mathcal{S}_0$, which necessitates the resolution of the equation $f_\theta = 0$, or the utilization of the marching cubes algorithm \cite{marchingcubes}. Both methods pose challenges in achieving fast and uniform point sampling. For the randomly sampled point $\bm{q}$ on non-zero level set $\mathcal{S}_{f_\theta(\bm{q})}$, we can also not filter this level set since there are no neighbor points on $\mathcal{S}_{f_\theta(\bm{q})}$.
    \item The normals utilized in our filtering are derived from the gradients of the neural network $f_\theta$. While the network typically offers reliable gradients, we may find that $\nabla f_\theta = 0$ is also the optimal solution to the minimum value of \cref{eq:weighted_project_distance,eq:dual_project_distance}. This degenerate solution is unexpected, as it implies a scenario where there is no surface when the gradient is zero everywhere.
\end{enumerate}
We will focus on addressing the two issues in the subsequent sections.

\begin{figure*}[t]
    \centering
    \includegraphics[width=\linewidth]{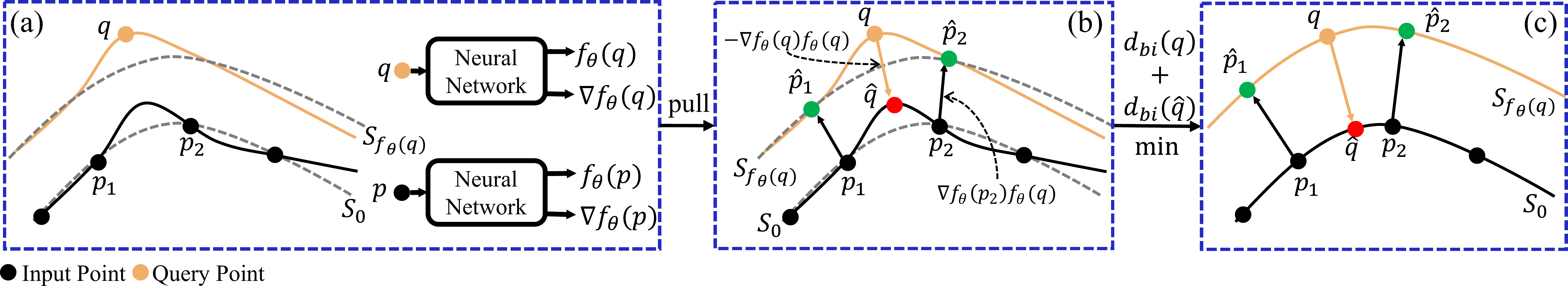}
    \caption{Overview of sampling points. (a) Sampling query points near the surface. (b) Pulling the query point to the zero level set and input points to the level set where the query point is located. (c) Applying the filter on each level set. The gradient $\nabla f_\theta$ in this figure defaults to be regularized.}
    \label{fig:fieldfiltering}
\end{figure*}
\subsubsection{Sampling points for filtering.}
Inspired by NeuralPull \cite{NPull}, we can pull a query point to the zero level set by the gradient of the neural network $f_\theta$. For a given query point $\bm{q}$ as input, the pulled location $\hat{\bm{q}}$ can be formulated as follows:

\begin{equation}
    \hat{\bm{q}} = \bm{q} - f_\theta(\bm{q}) \nabla f_\theta(\bm{q}) /  ||\nabla f_\theta(\bm{q})||.
    \label{Eq:npproject}
\end{equation}

The point $\bm{q}$ and $\hat{\bm{q}}$ lie respectively on level set $\mathcal{S}_{f_\theta(q)}$ and $\mathcal{S}_0$ as illustrate in \cref{fig:fieldfiltering}(b). By adopting the sampling strategy in NeuralPull, we can generate samples $\bm{Q}=\{\bm{q}_i|\bm{q}_i\in \bm{R}^3\}_{i=1}^M$ on different level sets near the surface and pull them to $\mathcal{S}_0$ by \cref{Eq:npproject}, to obtain $\hat{\bm{Q}}=\{\hat{\bm{q}}_i|\hat{\bm{q}}_i = \bm{q}_i - f_\theta(\bm{q}_i) \nabla f_\theta(\bm{q}_i) /  ||\nabla f_\theta(\bm{q}_i)||, \bm{q}_i \in \bm{Q}\}_{i=1}^M$. Hence, we can filter the zero level set by minimizing \cref{eq:dual_project_distance} across all pulled query points $\hat{\bm{Q}}$, which is equivalent to optimizing the following loss:

\begin{equation}
    % L_{zero} = \min\limits_\theta\sum_{\hat{\bm{q}}\in \hat{\bm{Q}}} d_{bi}(\hat{\bm{q}}).
    L_{zero} = \sum\nolimits_{\hat{\bm{q}}\in \hat{\bm{Q}}} d_{bi}(\hat{\bm{q}}),
    \label{eq:zerosetfilter}
\end{equation}
where for each $\hat{\bm{q}} \in \hat{\bm{Q}}$, $\mathcal{N}(\hat{\bm{q}}, \mathcal{S}_0)$ denotes finding the neighbors of $\hat{\bm{q}}$ within the input points $\bm{P}$, since $\bm{P}$ is assumed to be located on $\mathcal{S}_0$.

This filtering mechanism can be easily extended to non-zero level sets in a similar inverse manner. To be more specific, as for level set $S_{f_\theta(\bm{q})}$, the neighbor points for query point $\bm{q} \in \bm{Q}$ are required. These points should lie on the level set $S_{f_\theta(\bm{q})}$ same as $\bm{q}$, allowing us to filter the level set $S_{f_\theta(\bm{q})}$ using the same filter as described in \cref{eq:dual_project_distance}. 

However, obtaining $\mathcal{N}(\bm{q}, S_{f_\theta(q)})$ in $\bm{P}$ is not feasible, since all input points $\bm{P}$ are situated on the zero level set instead of the $S_{f_\theta(\bm{q})}$ level set. To address this issue, we propose a technique for identifying neighbors of $\bm{q}$ on level set $S_{f_\theta(\bm{q})}$, by projecting the input points $\bm{P}$ inversely onto the specific level set $S_{f_\theta(\bm{q})}$ based on the gradient, as depicted in \cref{fig:fieldfiltering}(b). The projected neighbor points can be represented as in \cref{eq:NP}. Filtering across multiple level sets helps to enhance the performance of our method by optimizing the consistency between different level sets within the SDF field, We further showcase this evidence in the ablation study detailed in Section \cref{ablation}.

\begin{equation}
\mathcal{N}(\bm{q}, S_{f_\theta(q)}) = \{\hat{\bm{p}} |\hat{\bm{p}} = \bm{p} + f_\theta(\bm{q}) \frac{\nabla f_\theta(\bm{p})}{||\nabla f_\theta(\bm{p})||}, \bm{p} \in \mathcal{N}(\hat{\bm{q}}, \mathcal{S}_0))\}.
\label{eq:NP}
\end{equation}

Based on the above analysis, we can filter the level sets $S_{f_\theta(\bm{q})}$ by minimizing \cref{eq:dual_project_distance} over all sample points $\bm{Q}$ through \cref{eq:NP}, equivalent to optimizing the following loss:
\begin{equation}
    % \min\limits_\theta\sum_{\bm{q}\in \bm{Q}} d(\bm{q}).
    L_{field} = \sum\nolimits_{\bm{q}\in \bm{Q}} d_{bi}(\bm{q}).
    \label{eq:levelsetfilter}
\end{equation}

\begin{wrapfigure}[9]{r}{0pt}
    \includegraphics[width=0.5\textwidth]{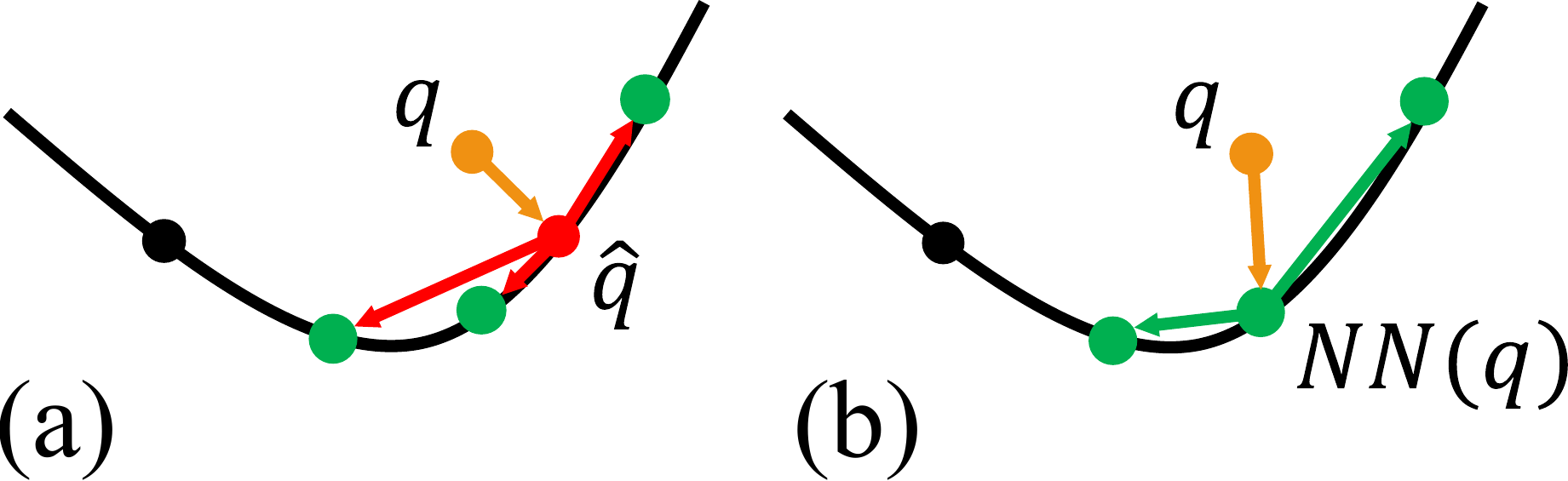}
    \caption{(a) Searching neighbors directly for $\hat{\bm{q}}$. 
    % (b) Substituting $\mathcal{N}(\hat{\bm{q}}, \mathcal{S}_0)$ with $\mathcal{N}(NN(\bm{q}), \mathcal{S}_0)$.
    (b) Searching neighbors for $NN(\bm{q})$ instead of $\hat{\bm{q}}$.
    }
    \label{fig:neighbor}
\end{wrapfigure}

It is worth noting that for a fixed query point $\bm{q}$, the pulled query point $\hat{\bm{q}}$ dynamically changes when training the neural network, which results in a time-consuming process to repeatedly conduct neighbor searching for $\hat{\bm{q}}$. To handle this matter, we substitute the $\mathcal{N}(\hat{\bm{q}}, \mathcal{S}_0)$ with $\mathcal{N}(NN(\bm{q}), \mathcal{S}_0)$, where $NN(\bm{q})$ denotes the nearest point of $\bm{q}$ within the point cloud $\bm{P}$ as shown in \cref{fig:neighbor}. While this substitution may introduce a slight bias for training, it also ensures the neighbor points are close to $\hat{\bm{q}}$, therefore this trade-off between efficiency and accuracy is reasonable.

% \subsubsection{Gradient constraint.}
\subsubsection{Gradient constraint.}
The other problem of implicit filtering is gradient degeneration. Overfitting the neural network requires the SDF to be geometrically initialized. We can consider the initialized implicit field as the noisy field and apply our filter directly to train the network from the beginning to fit the raw point cloud by removing the `noise'. However, if the denoise target is too complex, gradient degeneration will occur during the training process. Therefore, we need to add a constraint to the gradient of the SDF.

There are two ways for training the neural network to pull query points onto the surface based on NeuralPull \cite{NPull} and CAP-UDF \cite{CAP-UDF}. One is minimizing the distance between the pulled point $\hat{\bm{q}}$ and the nearest point $NN(\bm{q})$ as formulated below:

\begin{equation}
    L_{pull} = \frac{1}{M}\sum_{i \in [1, M]}{||\hat{\bm{q}}_i - NN(\bm{q}_i)||_2}.
\end{equation}

The other is minimizing the Chamfer distance between moved query points and the raw point cloud:

\begin{equation}
    L_{CD} = \frac{1}{M}\sum_{i \in [1, M]}{\min_{j \in [1, N]}{||\hat{\bm{q}}_i - \bm{p}_j||_2}} + \frac{1}{N}\sum_{j \in [1, N]}{\min_{i \in [1, M]}{||\bm{p}_j - \hat{\bm{q}}_i||_2}}.
\end{equation}

A stable SDF can be trained by the losses above since they are trying to move the query points to be in the same distribution with the point cloud, which can provide the constraint for our implicit filter. 
% However, when the optimization target is too complex, the neural network will likely fall into gradient degeneration. To prevent such a situation, it is necessary to employ either $L_{pull}$ or $L_{CD}$ to construct a stable signed distance field.
Here we choose $L_{CD}$ since the filtered points are likely not the nearest points and $L_{CD}$ is a more relaxed constraint.
% \cref{fig:2d} shows a 2D comparison of these losses, showing that our filter loss functions can reconstruct a field that is aligned at all level sets and maintains geometric characteristics. 

% \begin{figure*}[h]
%     \centering
%     \includegraphics[width=0.98\linewidth]{images/2d.pdf}
%     \caption{The 2D level sets show the distance field learned by different losses. The red lines represent the learned zero level set.}
%     \label{fig:2d}
% \end{figure*}

% \subsubsection{Loss function.}
\subsubsection{Loss function.}
Finally, our loss function is formulated as:
\begin{equation}
    L = L_{zero} + \alpha_1 L_{field} + \alpha_2 L_{dist} +  \alpha_3 L_{CD},
    \label{eq:loss}
\end{equation}
where $\alpha_1, \alpha_2$, and $\alpha_3$ is the balance weights for our implicit filtering loss.

% \subsubsection{Implementation details.}
\subsubsection{Implementation details.} We employ a neural network similar to OccNet \cite{OccupancyNetworks} and the geometric network initialization proposed in SAL\cite{SAL} with a smaller radius the same as GridPull\cite{chao2023gridpull} to learn the SDF. We use the strategy in NeuralPull\cite{NPull} to sample queries around each point $\bm{p}$ in $\bm{P}$. We set the weight $\alpha_3$ to 10 to constrain the learned SDF and $\alpha_1$ and $\alpha_2$ to 1. The parameters $\sigma_n, \sigma_p$ are set to $15^\circ, \max_{\bm{p}_j \in \mathcal{N}(\bm{\bar{p}}, \mathcal{S}_{f_\theta(\bar{p})})}(||\bar{\bm{p}} - \bm{p}_j||)$ respectively. 
% Instead of filtering the SDFs learned by other loss functions, we find our loss can directly learn the SDF from the raw point cloud even without $L_{CD}$.
% All the experiments are trained with our loss function from the beginning. In the experiment for scenes, we abort the $L_{field}$ since there exist open surfaces that cause the gradient orientation of the input points to be inconsistent.

\section{Experiments}
We conducted experiments to assess the performance of our implicit filter for surface reconstruction from raw point clouds. The results are presented for general shapes in \cref{sec:ExpShape}, real scanned raw data including 3D objects in \cref{sec:ExpReal}, and complex scenes in \cref{sec:ExpScene}. Additionally, ablation experiments were carried out to validate the theory and explore the impact of various parameters in \cref{ablation}.

\subsection{Surface Reconstruction for Shapes}
\label{sec:ExpShape}
\begin{wraptable}[13]{R}{0.63\textwidth}
  \centering
  \footnotesize
    \caption{Comparisons on ABC and Famous datasets. The threshold of F-score (F-S.) is 0.01.}
  \begin{tabular}{l|c |c | c| c| c| c}
    \hline
    \multirow{2}{*}{Methods} & \multicolumn{3}{c|}{ABC} & \multicolumn{3}{c}{FAMOUS} \\
    \cline{2-7}
   & $CD_{L2}$& $CD_{L1}$ & F-S. & $CD_{L2}$& $CD_{L1}$& F-S. \\
   
   \hline
    P2S\cite{Points2Surf}&0.298 &0.015&0.598 & 0.012 & 0.008 & 0.752\\
    IGR\cite{IGR}&2.675&0.063&0.448&1.474&0.044&0.573\\
    NP\cite{NPull} &0.095  &0.011& 0.673 &0.100& 0.012 & 0.746 \\
    PCP\cite{PredictiveContextPriors} &0.252& 0.023&0.373 &0.037& 0.014&0.435\\
    SIREN\cite{siren}&0.022&0.012& 0.493&0.025&0.012&0.561 \\
     DIGS\cite{ben2022digs}&0.021&0.010&0.667 & 0.015&0.008 & 0.772\\
    
    \hline
    Ours& \textbf{0.011}& \textbf{0.009}& \textbf{0.691} & \textbf{0.008} & \textbf{0.007} & \textbf{0.778}\\
  \hline
  \end{tabular}
  \label{tab:abc}
\end{wraptable}
\subsubsection{Datasets and metrics.}
For surface reconstruction of general shapes from raw point clouds, we conduct evaluations on three widely used datasets including a subset of ShapeNet\cite{shapenet}, ABC\cite{abc}, and FAMOUS\cite{Points2Surf}. We use the same setting with NeuralPull\cite{NPull} for the dataset ShapeNet. For datasets ABC and FAMOUS, we use the train/test splitting released by Points2Surf\cite{Points2Surf} and we sample points directly from the mesh in the ABC dataset without other mesh preprocessing to keep the sharp features. 

\begin{figure*}[t]
    \centering
    \includegraphics[width=0.98\linewidth]{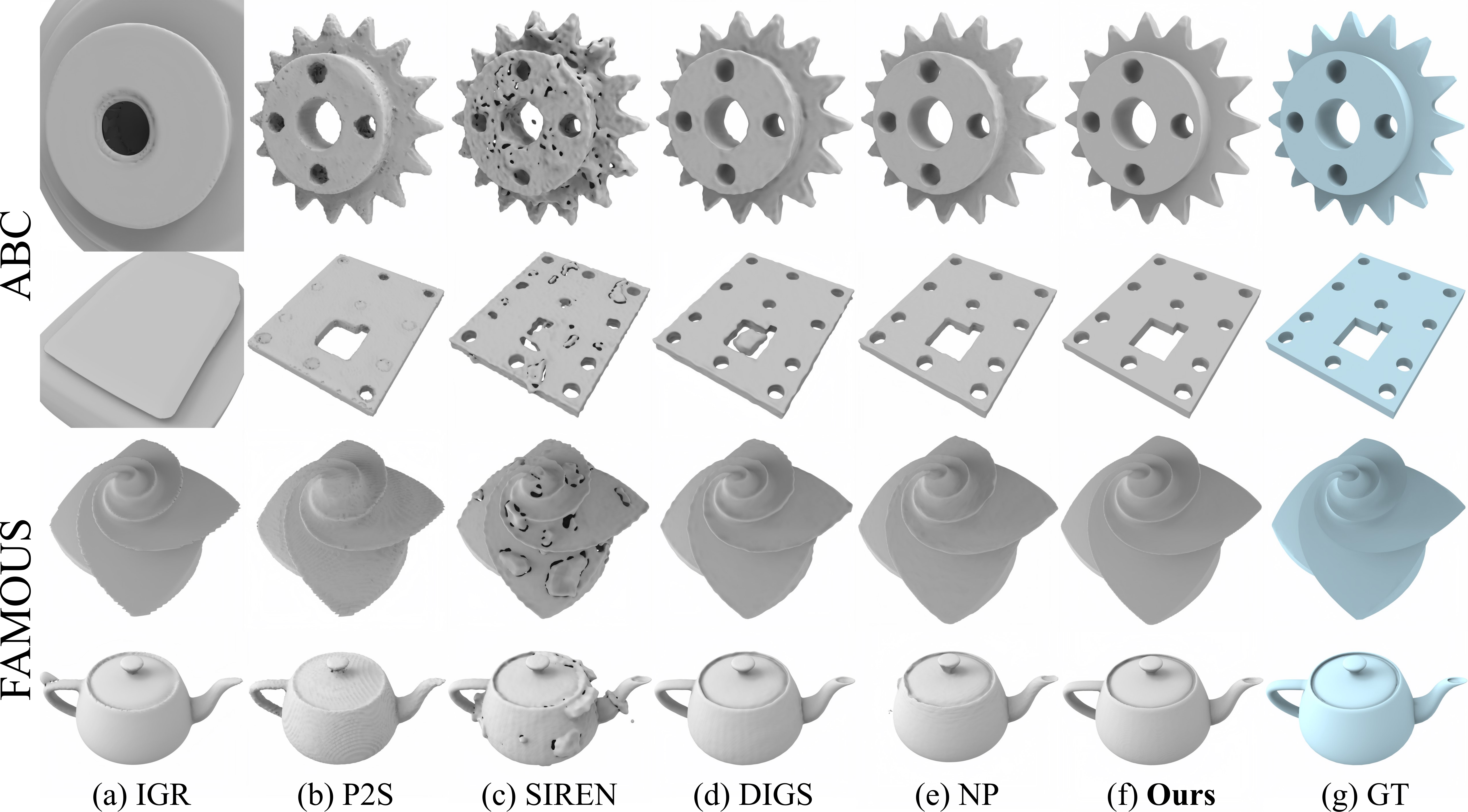}
    \caption{Visual comparisons of surface reconstruction on ABC and FAMOUS datasets. Our method can reconstruct objects with sharp edges and less noise compared with other methods.}
    \label{fig:abc}
\end{figure*}

For evaluating the performance, we follow NeuralPull to sample $1\times10^5$ points from the reconstructed surfaces and the ground truth meshes on the ShapeNet dataset and sample $1\times10^4$ on the ABC and FAMOUS datasets. For the evaluation metrics, we use L1 and L2 Chamfer distance ($CD_{L1}$ and $CD_{L2}$) to measure the error. Moreover, we adopt normal consistency (NC) and F-score to evaluate the accuracy of the reconstructed surface, the threshold is the same with NeuralPull.

\subsubsection{Comparisons.}
\begin{wrapfigure}[14]{R}{0.5\textwidth}
\centering
    % \raisebox{0pt}[\dimexpr\height-1\baselineskip\relax]{
    \includegraphics[width=0.46\textwidth]{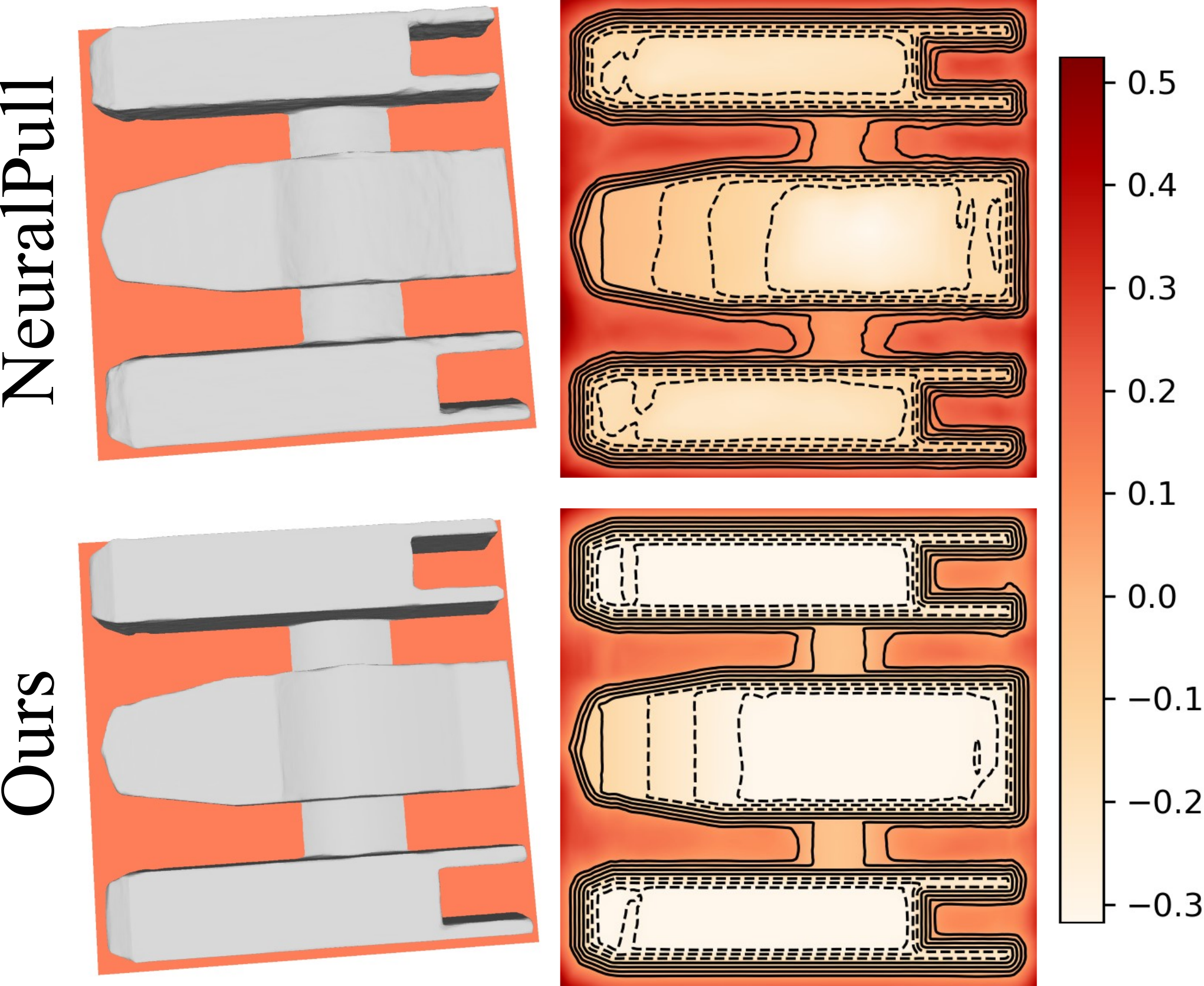}
    % }
    \caption{Visualization of level sets on a cross section.}
    \label{fig:level_set}
\end{wrapfigure}
\begin{figure*}[t]
    \centering
    \includegraphics[width=0.98\linewidth]{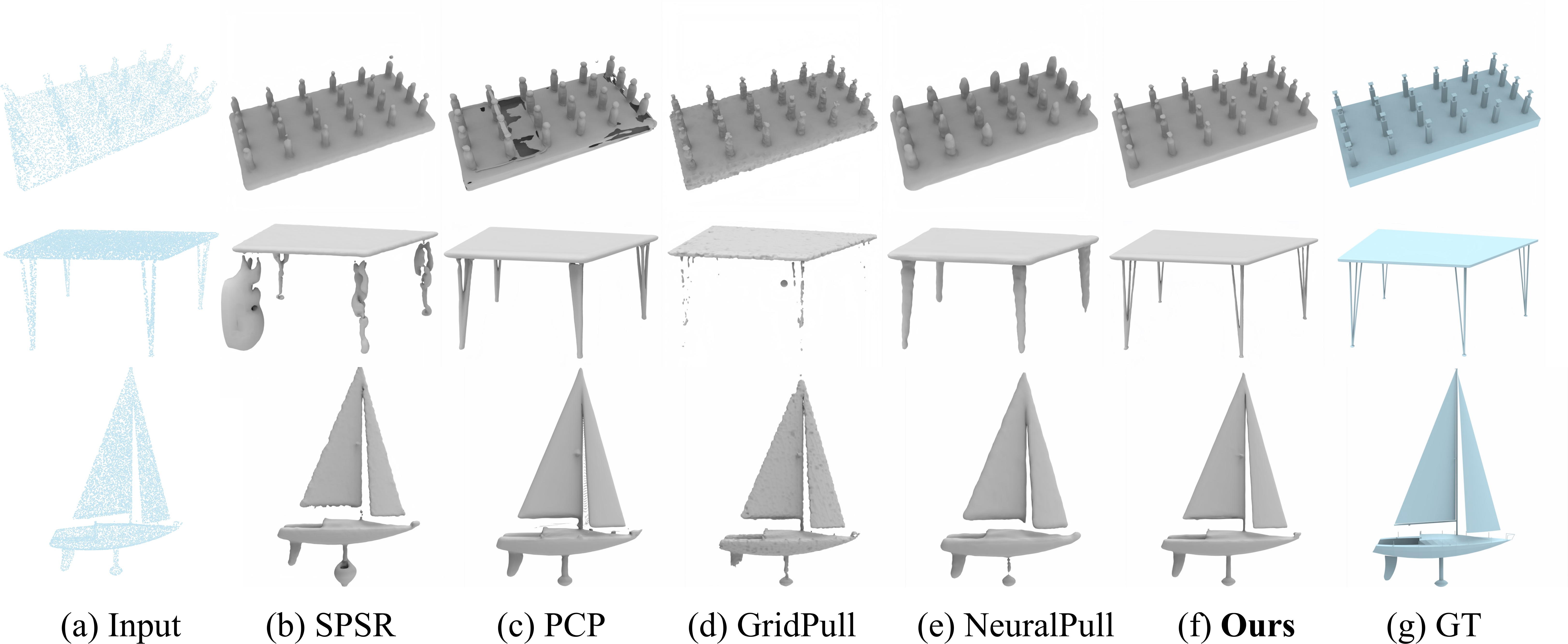}
    \caption{Visual comparisons of surface reconstruction on ShapeNet dataset.}
    \label{fig:shapenet}
\end{figure*}

To evaluate the validity of our implicit filter, we compare our method with a variety of methods including SPSR\cite{kazhdan2013screened}, Points2Surf (P2S)\cite{Points2Surf}, IGR\cite{IGR}, NeuralPull (NP)\cite{NPull}, LPI\cite{LPI}, PCP\cite{PredictiveContextPriors}, GridPull (GP)\cite{chao2023gridpull}, SIREN\cite{siren}, DIGS\cite{ben2022digs}. The quantitative results on ABC and FAMOUS datasets are shown in \cref{tab:abc}, and selectively visualized in \cref{fig:abc}. Our model reaches state-of-the-art performance on both datasets, accomplishing the goal of eliminating noise on each level set while preserving the geometric details. 
To more intuitively validate the efficacy of our filtering, we visualize the level sets on a cross section in \cref{fig:level_set}. We also report the results on ShapeNet which contains over 3000 objects in terms of $CD_{L2}$, NC, and F-Score with thresholds of 0.002 and 0.004 in \cref{tab:shapenet}. The detailed comparison for each class of ShapeNet can be found in the supplementary material. Our method outperforms previous methods over most classes. The visualization comparisons in \cref{fig:shapenet} show that our method can reconstruct a smoother surface with fine details.

\begin{table}[t]
  \centering
    \caption{Comparisons on ShapeNet dataset.}
\begin{tabular}{ l | c | c | c | c | c | c }
    \hline
     &SPSR\cite{kazhdan2013screened} & NP \cite{NPull} &LPI\cite{LPI} &PCP\cite{PredictiveContextPriors}&GP
 \cite{chao2023gridpull}&Ours  \\
    \hline
    $CD_{L2} \times 100$ & 0.286 & 0.038& 0.0171 & 0.0136& 0.0086& \textbf{0.0032}\\
        NC & 0.866 & 0.939& 0.9596 & 0.9590& 0.9723& \textbf{0.9779}\\
        F-Score (0.002) & 0.407 & 0.961& 0.9912 & 0.9871& 0.9896& \textbf{0.9976}\\
        F-Score (0.004)&0.618&0.976&0.9957&0.9899&0.9923&\textbf{0.9985}\\
  \hline
  \end{tabular}
  \label{tab:shapenet}
\end{table}

\begin{figure*}[ht]
    \centering 
    \includegraphics[width=0.98\linewidth]{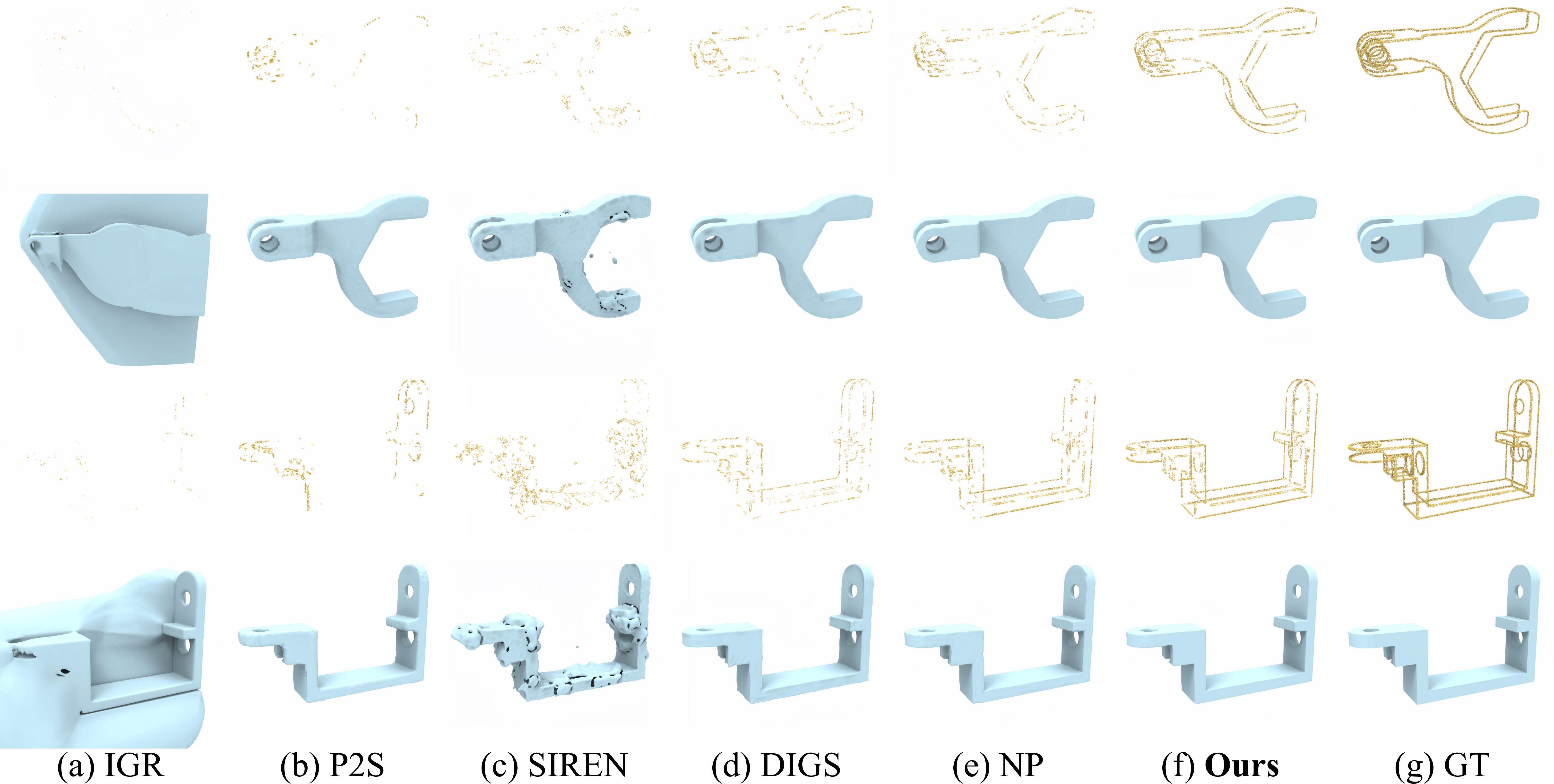}
    \caption{Visual comparisons of edge points and reconstruction results.}
    \label{fig:abl_edge}
\end{figure*}

To validate the effect of our filter on sharp geometric features. We evaluate the edge points by the edge Chamfer distance metric used in \cite{chen2020bspnet}. We sample 100k points uniformly on the surface of both the reconstructed mesh and ground truth.
% to calculate the edge points. The results are shown in \cref{tab:abl_ecd} and visualized in \cref{fig:abl_edge}.
The edge point $\bm{p}$ is calculated by finding whether there exists a point $\bm{q} \in \mathcal{N}_\epsilon(\bm{p})$ satisfied $|\bm{n}_q\bm{n}_p| < \sigma$, where $\mathcal{N}_\epsilon(\bm{p})$ represents the neighbor points within distance $\epsilon$ from $\bm{p}$. The results are shown in \cref{tab:abl_ecd} and visualized in \cref{fig:abl_edge}. We set $\epsilon = 0.01$ and $\sigma = 0.1$.

\begin{table}[t]
\centering
    \caption{Edge Chamfer distance comparisons on ABC dataset, $ECD_{L2} \times 100$.}
    \begin{tabular}{ c  |c   |c  |c |c |c|c|c}
    \hline
    Methods & P2S\cite{Points2Surf} & IGR\cite{IGR} & NP\cite{NPull} & PCP\cite{PredictiveContextPriors} & SIREN\cite{siren} & DIGS\cite{ben2022digs} & Ours \\
   \hline
   $ECD_{L1}$ & 0.0496& 0.0835& 0.0501& 0.0628& 0.0695& 0.0786& \textbf{0.0256} \\
   $ECD_{L2}$ & 1.055& 2.365& 1.255& 1.265& 1.407& 2.493& \textbf{0.399}\\
   \hline
  \end{tabular}
  \label{tab:abl_ecd}
\end{table}

\subsection{Surface Reconstruction for Real Scans}
\label{sec:ExpReal}
\subsubsection{Dataset and metrics.} For surface reconstruction of real point cloud scans, we follow VisCo\cite{visco} to evaluate our method under the Surface Reconstruction Benchmarks (SRB)\cite{SRB}. We use Chamfer and Hausdorff distances ($CD_{L1}$ and HD) between the reconstruction meshes and the ground truth. Furthermore, we report their corresponding one-sided distances ($d_{\overrightarrow{C}}$ and $d_{\overrightarrow{H}}$) between the reconstructed meshes and the input noisy point cloud.

\begin{wrapfigure}[9]{r}{0pt}
    % \raisebox{0pt}[\dimexpr\height-2.5\baselineskip\relax]{
    \includegraphics[width=0.6\textwidth]{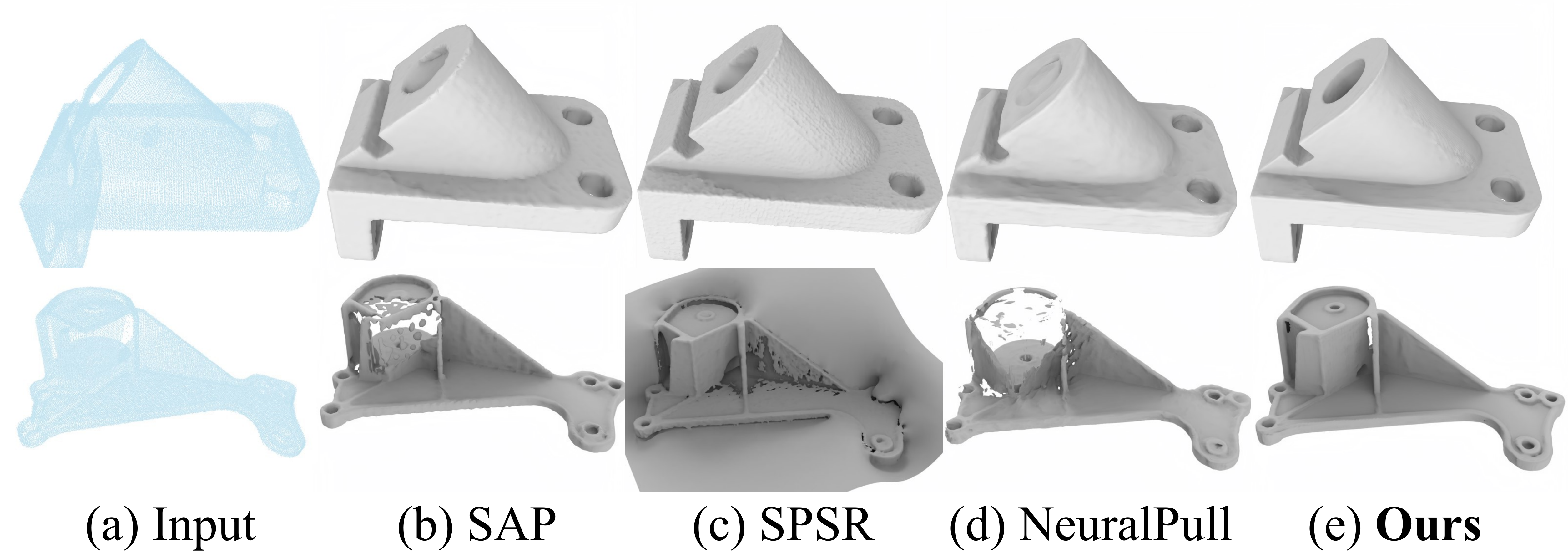}
    % }%
    \caption{Visual comparisons on SRB dataset.}
    \label{fig:srb}
\end{wrapfigure}
\subsubsection{Comparisons.} We compare our method with state-of-the-art methods under the real scanned SRB dataset, including IGR\cite{IGR}, SPSR\cite{kazhdan2013screened}, Shape As Points (SAP)\cite{SAP}, NeuralPull (NP)\cite{NPull}, and GridPull (GP)\cite{chao2023gridpull}. The numerical comparisons are shown in \cref{tab:srb}, where we achieve the best accuracy in most cases. The visual comparisons in \cref{fig:srb} demonstrate that our method can reconstruct a continuous and smooth surface with geometry details.

\begin{table}[h]
  \caption{Comparisons on SRB dataset.}
  \label{tab:srb}
  \centering
  \footnotesize
\resizebox{\textwidth}{!}{
  \begin{tabular}{ c  |c  |c  |c  |c  |c  |c  |c  |c  | c | c}
  % \begin{tabularx}{0.9\textwidth}{@{}l|c|c|c|c|c|c@{}}
    \hline
   &&SPSR\cite{kazhdan2013screened}&IGR\cite{IGR}&SIREN\cite{siren}&VisCo\cite{visco}&SAP\cite{SAP} & NP\cite{NPull} &GP\cite{chao2023gridpull}  &DIGS \cite{ben2022digs}&Ours  \\
    \hline
    \multirow{4}{*}{Anchor} 
    &$CD_{L1}$ & 0.60 & 0.22& 0.32 & 0.21 & 0.12&  0.122
&0.093 &0.063&\textbf{0.052}\\
    &HD & 14.89 & 4.71& 8.19 & 3.00& 2.38&  3.243
&1.804 &1.447&\textbf{1.232}\\
    &$d_{\overrightarrow{C}}$ & 0.60 & 0.12& 0.10 & 0.15  & 0.08 &  0.061
&0.066 &0.030&\textbf{0.025}\\
    &$d_{\overrightarrow{H}}$& 14.89 & 1.32 & 2.432 & 1.07& 0.83&  3.208
&0.460  &0.270&\textbf{0.265}\\
    \hline
    \multirow{4}{*}{Daratech} 
    &$CD_{L1}$ & 0.44& 0.25& 0.21 & 0.21 & 0.26&  0.375
&0.062 &0.049&\textbf{0.051}\\
    &HD & 7.24 & 4.01& 4.30 & 4.06& 0.87&  3.127
&\textbf{0.648} &0.858&0.751\\
    &$d_{\overrightarrow{C}}$& 0.44 & 0.08& 0.09 & 0.14  & 0.04 &  0.746
&0.039 &0.025&\textbf{0.028}\\
    &$d_{\overrightarrow{H}}$ & 7.24 & 1.59& 1.77 & 1.76& 0.41&   3.267
&\textbf{0.293} &0.441&0.423\\
        \hline
    \multirow{4}{*}{DC} 
    &$CD_{L1}$ & 0.27 & 0.17&0.15 & 0.15 & 0.07&  0.157
&0.066 &0.042&\textbf{0.041}\\
    &HD & 3.10 & 2.22&2.18 & 2.22& 1.17&  3.541
&1.103 &\textbf{0.667}&0.815\\
    &$d_{\overrightarrow{C}}$ & 0.27&0.09&0.06&0.09&0.04& 0.242
&0.036  &0.022&\textbf{0.019}\\
    &$d_{\overrightarrow{H}}$ & 3.10&2.61&2.76&2.76&\textbf{0.53}& 3.523
&0.539 &0.729&\textbf{0.724}\\
    \hline
    \multirow{4}{*}{Gargoyle} 
    &$CD_{L1}$ & 0.26 &0.16&0.17 & 0.17 & 0.07&  0.080
&0.063 &0.047&\textbf{0.044}\\
    &HD & 6.80 & 3.52& 4.64 & 4.40& 1.49&  1.376
&1.129 &\textbf{0.971}&1.089\\
    &$d_{\overrightarrow{C}}$ & 0.26&0.06&0.08&0.11&0.05& 0.063
&0.045 &0.028&\textbf{0.022}\\
    &$d_{\overrightarrow{H}}$ &6.80&0.81&0.91&0.96&0.78& 0.475
&0.700 &0.271&\textbf{0.246}\\
        \hline
    \multirow{4}{*}{Lord Quas} 
    &$CD_{L1}$ & 0.20 & 0.12& 0.17 & 0.12  & 0.05 &  0.064
&0.047 &0.031&\textbf{0.030}\\
    &HD & 4.61 & 1.17& 0.82 & 1.06& 0.98&  0.822
&0.569 &\textbf{0.496}&0.554\\
    &$d_{\overrightarrow{C}}$ & 0.20&0.07&0.12&0.07&0.04& 0.053
&0.031 &0.017&\textbf{0.014}\\
    &$d_{\overrightarrow{H}}$ &4.61&0.98&0.76&0.64&0.51& 0.508&0.370 &\textbf{0.181}&0.230\\
    \hline
  \end{tabular}
}
\end{table}

\subsection{Surface Reconstruction for Scenes}
\label{sec:ExpScene}
\subsubsection{Dataset and metrics.} To further demonstrate the advantage of our method in the surface reconstruction of real scene scans, we conduct experiments using the 3D Scene dataset. The 3D Scene dataset is a challenging real-world dataset with complex topology and noisy open surfaces. We uniformly sample 1000 points per $m^2$ of each scene as the input and follow PCP\cite{PredictiveContextPriors} to sample 1M points on both the reconstructed and the ground truth surfaces. We leverage L1 and L2 Chamfer distance ($CD_{L1}, CD_{L2}$) and normal consistency (NC) to evaluate the reconstruction quality.

\subsubsection{Comparisons.} 
\begin{figure*}[t]
    \centering
    \includegraphics[width=0.98\linewidth]{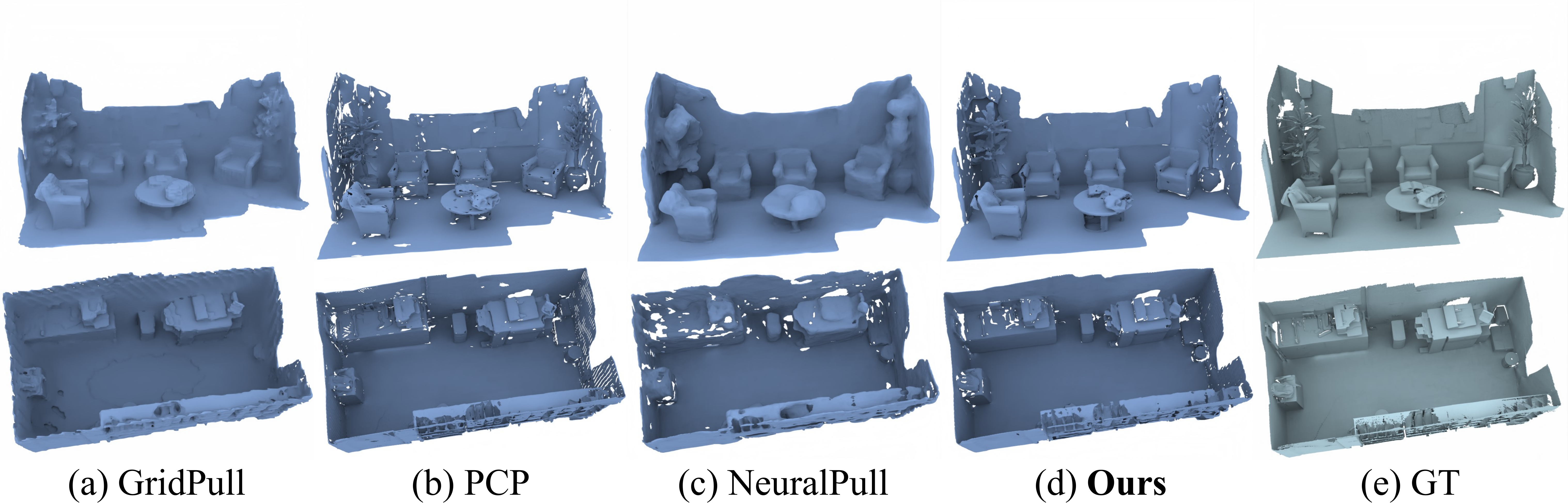}
    \caption{Visual comparisons of surface reconstruction on 3D Scene dataset.}
    \label{fig:3dscene}
\end{figure*}

We compare our method with the state-of-the-art methods ConvONet\cite{convonet}, LIG\cite{LIG}, DeepLS\cite{Deeplocalshape}, NeuralPull (NP)\cite{NPull}, PCP\cite{PredictiveContextPriors}, GridPull (GP)\cite{chao2023gridpull}. The numerical comparisons in \cref{tab:3dscene} demonstrate our superior performance in all scenes even compared with the local-based methods. We further present visual comparisons in \cref{fig:3dscene}. The visualization further shows that our method can achieve smoother with high-fidelity surfaces in complex scenes. It should be noted that the surface we extract here is not the zero level set but the 0.001 level set since the scene is not watertight. For NeuralPull we use the threshold of 0.005 instead of 0.001 to extract the complete surface therefore the mesh looks thicker.

\begin{table}[h]
  \caption{Comparisons on 3D Scene dataset, $CD_{L2} \times 1000$.}
  \label{tab:3dscene}
  \centering
  \footnotesize
  \resizebox{\textwidth}{!}{
  \begin{tabular}{ l | c | c | c | c | c | c | c| c| c| c| c| c| c| c| c}
  % \begin{tabularx}{0.9\textwidth}{@{}l|c|c|c|c|c|c@{}} c
    \hline
     & \multicolumn{3}{c|}{Burghers} & \multicolumn{3}{c|}{Lounge} &\multicolumn{3}{c|}{Copyroom}&\multicolumn{3}{c|}{Stonewall}&\multicolumn{3}{c}{Totempole} \\
    \hline
    & $CD_{L2}$ & $CD_{L1}$& NC & $CD_{L2}$ & $CD_{L1}$& NC & $CD_{L2}$ & $CD_{L1}$& NC & $CD_{L2}$& $CD_{L1}$ & NC & $CD_{L2}$ & $CD_{L1}$& NC \\
    \hline
    ConvONet\cite{convonet}&27.46&0.079&0.907&9.54&0.046&0.894&10.97&0.045&0.892&20.46&0.069&0.905&2.054&0.021&0.943\\
    LIG\cite{LIG}&3.055&0.045&0.835&9.672&0.056&0.833&3.61&0.036&0.810&5.032&0.042&0.879&9.58&0.062&0.887 \\
    DeepLS\cite{Deeplocalshape}&0.401&0.017&0.920&6.103&0.053&0.848&0.609&0.021&0.901&0.320&0.015&0.954&0.601&0.017&0.950\\
    % PCP \cite{PredictiveContextPriors}&0.189&0.936&0.102&0.945&0.116&0.939&0.146&0.965&0.395&0.958 \\
    GP\cite{chao2023gridpull}&1.367&0.028&0.873&4.684&0.053&0.827&2.327&0.030&0.857&2.234&0.024&0.913&2.278&0.034&0.878 \\
    PCP\cite{PredictiveContextPriors}&1.339&0.031&0.929&0.432&0.014&\textbf{0.934}&0.405&0.014&\textbf{0.914}&0.266&0.014&\textbf{0.957}&1.089&0.029&\textbf{0.954}\\
    % NP\cite{NPull}& 0.357&0.014&0.907&1.022&0.015&0.899&0.360&0.012&0.864&0.085&\textbf{0.008}&0.950&\textbf{0.197}&\textbf{0.013}&0.946\\
    NP\cite{NPull}&0.897&0.025&0.883&0.855&0.022&0.887&0.479&0.018&0.862&0.434&0.018&0.929&1.604&0.032&0.923\\
    \hline
    % Ours& \textbf{0.186}&\textbf{0.938}&0.103&\textbf{0.945}&0.119&0.936&\textbf{0.137}&\textbf{0.966}&\textbf{0.328}&\textbf{0.960}\\
        % Ours& \textbf{0.133}&\textbf{0.933}&\textbf{0.176}&0.931&\textbf{0.112}&0.913&\textbf{0.082}&\textbf{0.957}&0.202&0.944\\
    Ours&\textbf{0.133}&\textbf{0.011}&\textbf{0.934}&\textbf{0.120}&\textbf{0.008}&0.926&\textbf{0.111}&\textbf{0.009}&0.913&\textbf{0.082}&\textbf{0.009}&\textbf{0.957}&\textbf{0.203}&\textbf{0.013}&0.944 \\
  \hline
  \end{tabular}
  }
  \end{table}

\subsection{Ablation Studies}
\label{ablation}

We conduct ablation studies on the FAMOUS dataset to demonstrate the effectiveness of our proposed implicit filter and explore the effect of some important hyperparameters. We report the performance in terms of L1 and L2 Chamfer distance ($CD_{L1}, CD_{L2}\times 10^3$), normal consistency (NC), and F-Score (F-S.).

\begin{wraptable}[6]{r}{0.6\textwidth}
\captionsetup{aboveskip=0pt}
    \caption{Effect of the Eikonal term.}
\centering
\footnotesize
  \begin{tabular}{ c  |c   |c  |c |c }
    \hline
Loss & $CD_{L1}$ & $CD_{L2}$ & F-S. & NC \\
   \hline
   w/ Eikonal, w/o CD &0.009 & 0.021 & 0.738 & 0.899\\
   w/ Eikonal, w/ CD & 0.008 & 0.009 & 0.774 & 0.910 \\
   w/o Eikonal, w/ CD & \textbf{0.007} & \textbf{0.008} & \textbf{0.778} & \textbf{0.911}\\
   \hline
  \end{tabular}
  \label{tab:eik}
\end{wraptable}
\noindent \textbf{Effect of Eikonal loss.}
We select the $L_{CD}$ to prevent the degeneration of the gradient since it both constrains the value and the gradient of the SDF. It also guides how to pull the query point onto the surface. Therefore we omit the Eikonal term used in previous methods like the IGR\cite{IGR}, SIREN\cite{siren}, and DIGS\cite{ben2022digs} which have no other direct supervision for the gradient. To verify this selection, we conduct the following experiments by trade-off these two functions. With the experimental results in \cref{tab:eik}, we find that only applying the Eikonal term is not as effective as CD alone. At the same time combining the Eikonal term with CD does not further enhance the experiment results, but the difference is small.

\begin{figure*}[t]
    \centering
    \includegraphics[width=0.98\linewidth]{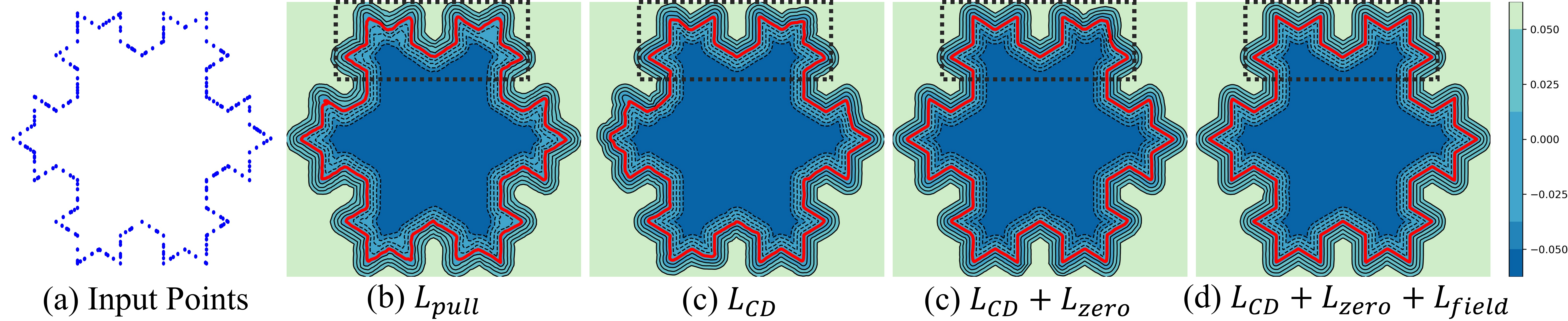}
    \caption{The 2D level sets show the distance field learned by different losses. The red lines represent the learned zero level set.}
    \label{fig:2d}
\end{figure*}

\noindent \textbf{Effect of level set filtering.}
To justify the effectiveness of each term in our loss function. We report the results trained by different combinations in \cref{tab:abl_loss}. The $L_{CD}$ is more applicable for training SDF from raw point clouds. The zero-level filter can help remove the noise and keep the geometric features. 
Filtering across non-zero level sets can improve the overall consistency of the entire signed distance field. Since we assume all input points lie on the surface, the function $L_{dist}$ is also necessary. \cref{fig:2d} shows a 2D comparison of these losses, showing that our filter loss functions can reconstruct a field that is aligned at all level sets and maintains geometric characteristics.

\noindent \textbf{Effect of the bidirectional projection.} 
To validate our bidirectional projection distance, we report the results in \cref{tab:abl_bidire}. The numerical comparisons show that projecting the distance to both normals can improve the reconstruction quality. Note that only using $d(\bm{\bar{p}})$ can also improve the results.

\begin{table}[h]
% \centering%
\begin{minipage}[b]{0.6\textwidth}
% \centering
\small
\caption{Effect of different losses.}
  \begin{tabular}{ c  |c   |c  |c |c }
    \hline
   Loss & $CD_{L1}$ & $CD_{L2}$ & F-S. & NC \\
   \hline
   $L_{pull}$ & 0.012 & 0.083 & 0.742 & 0.884 \\
   $L_{CD}$ &0.010 & 0.031 & 0.757 & 0.891 \\
   $L_{CD} + L_{zero}$ & 0.008 & 0.018 & 0.772 & 0.905 \\
   $L_{CD} + L_{zero} + L_{field}$ & 0.008 & 0.011 & 0.769 & 0.908 \\
   Ours & \textbf{0.007} & \textbf{0.008} & \textbf{0.778} & \textbf{0.911}\\
   \hline
  \end{tabular}
  \label{tab:abl_loss}
\end{minipage}
\begin{minipage}[b]{0.35\textwidth}
\centering
\small
\caption{Effect of bidirectional projection.}
  \begin{tabular}{ c  |c   |c }
    \hline
    &$d(\bm{\bar{p}})$&   $d_{bi}(\bm{\bar{p}})$ \\
    
   \hline
 $CD_{L1}$ &  0.010 & \textbf{0.007}\\
 $CD_{L2}$& 0.024& \textbf{0.008}\\
 F-S. & 0.726 & \textbf{0.778}\\
 NC & 0.890 & \textbf{0.911}\\
   \hline
  \end{tabular}
  \label{tab:abl_bidire}
\end{minipage}
\end{table}

\noindent \textbf{Weight of level set projection loss.} We explore the effect of the $L_{CD}$ loss function by adjusting the weight $\alpha_3$ in  \cref{eq:loss}. We report our results with different candidates \{0, 1, 10\} in  \cref{tab:abl_alpha}, where 0 means we do not use the $L_{CD}$ to constrain the gradient. The comparisons in \cref{tab:abl_alpha} show that although our implicit filter can directly learn SDFs, it is better to adopt the $L_{CD}$ for a more stable field. However, if the weight is too large, the filtering effect will decrease. It is recommended to select weights ranging from 1 to 10, which is usually adequate. For the weights $\alpha_1$ and $\alpha_2$, setting them to 1 is always necessary.

\noindent \textbf{Effect of filter parameters.}
We compare the effect of different parameters $\sigma_n, \sigma_p$ in \cref{tab:abl_filter_para}. The diagonal weight for $\sigma_p$ means the length of the diagonal of the bounding box for the local patch mentioned in \cite{pointfilter}. The results indicate that the method is relatively robust to parameter variation in a certain range.

\begin{table}[h]
% \centering%
\begin{minipage}[b]{0.43\textwidth}
\centering%
    \caption{Effect of weight $\alpha_3$.}
  \begin{tabular}{ c  |c   |c  |c |c }
    \hline
$\alpha_3$ & $CD_{L1}$ & $CD_{L2}$ & F-S. & NC \\
   \hline
   0 & 0.008 & 0.013 & 0.758 & 0.903 \\
   1 & \textbf{0.007} & 0.011 & 0.772 & 0.910 \\
   10 & \textbf{0.007} & \textbf{0.008} & \textbf{0.778} & \textbf{0.911} \\
   100 & 0.008 & 0.009 & 0.774 & 0.909 \\
   \hline
  \end{tabular}
  \label{tab:abl_alpha}
\end{minipage}
\begin{minipage}[b]{0.55\textwidth}
\centering%
    \caption{Effect of filter parameters $\sigma_n$ and $\sigma_p$.}
  \begin{tabular}{ c| c  |c   |c  |c |c }
    \hline
  &  & $CD_{L1}$ & $CD_{L2}$ & F-S. & NC \\
   \hline
  \multirow{4}{*}{$\sigma_n$} 
  & $15^\circ$ & \textbf{0.007} & \textbf{0.008} & \textbf{0.778} & \textbf{0.911} \\
  & $30^\circ$ & \textbf{0.007} & 0.011 & 0.771 & 0.907 \\
  & $45^\circ$ & 0.008 & 0.012 & 0.764 & 0.903 \\
  & $60^\circ$ & 0.008 & 0.010 & 0.767 & 0.901 \\
   \hline
  \multirow{2}{*}{$\sigma_p$} 
  & max & \textbf{0.007} & \textbf{0.008} & \textbf{0.778} & \textbf{0.911} \\
  & diagonal &0.008 & 0.011 & 0.763 & 0.904\\
  \hline
  \end{tabular}
  \label{tab:abl_filter_para}
\end{minipage}
\end{table}

\section{Conclusion}
We introduce implicit filtering on SDFs to reduce the noise of the signed distance field while preserving geometry features. We filter the distance field by minimizing the weighted bidirectional projection distance, where we can generate sampling points on the zero level set and neighbor points on non-zero level sets by the pulling procedure. By leveraging the Chamfer distance, we address the issue of gradient degeneration problem. The visual and numerical comparisons demonstrate our effectiveness and superiority over state-of-the-art methods.

% \section{Conclusion}
% The paper ends with a conclusion. 

% \clearpage\mbox{}Page \thepage\ of the manuscript.
% \clearpage\mbox{}Page \thepage\ of the manuscript.
% \clearpage\mbox{}Page \thepage\ of the manuscript.
% \clearpage\mbox{}Page \thepage\ of the manuscript.
% \clearpage\mbox{}Page \thepage\ of the manuscript. This is the last page.
% \par\vfill\par
% Now we have reached the maximum length of an ECCV \ECCVyear{} submission (excluding references and acknowledgements).
% References should start immediately after the main text, but can continue past p.\ 14 if needed. 
% \clearpage  % TODO FINAL: This \clearpage needs to be removed from both review and camera-ready versions.

\section*{Acknowledgements}
The corresponding author is Ge Gao. This work was supported by Beijing Science and Technology Program (Z231100001723014).

% ---- Bibliography ----
%
% BibTeX users should specify bibliography style 'splncs04'.
% References will then be sorted and formatted in the correct style.
%

\bibliographystyle{splncs04}
\bibliography{main}
\end{document}